\documentclass[10pt,twocolumn,letterpaper]{article}
\usepackage{cvpr}
\usepackage{times}
\usepackage{epsfig}
\usepackage{graphicx}
\usepackage{amsmath}
\usepackage{amssymb}
\usepackage{multirow}
\usepackage{subcaption}
\usepackage[pagebackref=true,breaklinks=true,letterpaper=true,colorlinks,bookmarks=false]{hyperref}

\cvprfinalcopy % *** Uncomment this line for the final submission

\def\cvprPaperID{4943} % *** Enter the CVPR Paper ID here

\ifcvprfinal\fi
\begin{document}

%%%%%%%%% TITLE
\title{VideoSSL: Semi-Supervised Learning for Video Classification}

\author{
  Longlong Jing${^{1}}$ \thanks{The work was partially done at Comcast Applied AI Research, Washington, DC.} \quad Toufiq Parag${^2}$ \quad Zhe Wu${^2}$ \quad Yingli Tian${^1}$ \quad Hongcheng Wang${^2}$\\ 
  $^1$The City University of New York, ~$^2$Comcast Applied AI Research\\
%   {\tt\small $^1$ $\{$secondauthor$\}$@i2.org, $^2$ $\{$}
}

% \and

\maketitle
%\thispagestyle{empty}

%%%%%%%%% ABSTRACT
\begin{abstract}
We propose a semi-supervised learning approach for video classification, VideoSSL, using convolutional neural networks (CNN). Like other computer vision tasks, existing supervised video classification methods demand a large amount of labeled data to attain good performance. However, annotation of a large dataset is expensive and time consuming. To minimize the dependence on a large annotated dataset, our proposed semi-supervised method trains from a small number of labeled examples and exploits two regulatory signals from unlabeled data. The first signal is the pseudo-labels of  unlabeled examples computed from the confidences of the CNN being trained. The other is the normalized probabilities, as predicted by an image classifier CNN, that captures the information about appearances of the interesting objects in the video. We show that, under the supervision of these guiding signals from unlabeled examples, a video classification CNN can achieve impressive performances utilizing a small fraction of annotated examples on three publicly available datasets: UCF101, HMDB51 and Kinetics.

\end{abstract}

%%%%%%%%% BODY TEXT
\section{Introduction}

Video understanding has been a topic of interest in computer vision community for many years. Although video understanding and analytics tasks such as action recognition have been pioneered by early classical vision studies \cite{laptev05interest,niebles06bmvc}, the more recent methods have gained much success with CNNs \cite{C3D, 2Plus1D}. Among many CNN based algorithms for video classification exploiting different type of information extracted from the video (RGB values or optical flow) and various network architectures (Two stream \cite{Two-Stream}, LSTM \cite{LRCN, LSTM, Beyond}, 3D CNN \cite{3D}), the variants of 3D CNNs utilizing the spatiotemporal features have produced the state of the art results \cite{C3D, 3DResNet, P3D, I3D, 2Plus1D, gonda18bmvc}.  

\begin{figure}
\begin{center}
\includegraphics[width=0.45\textwidth]{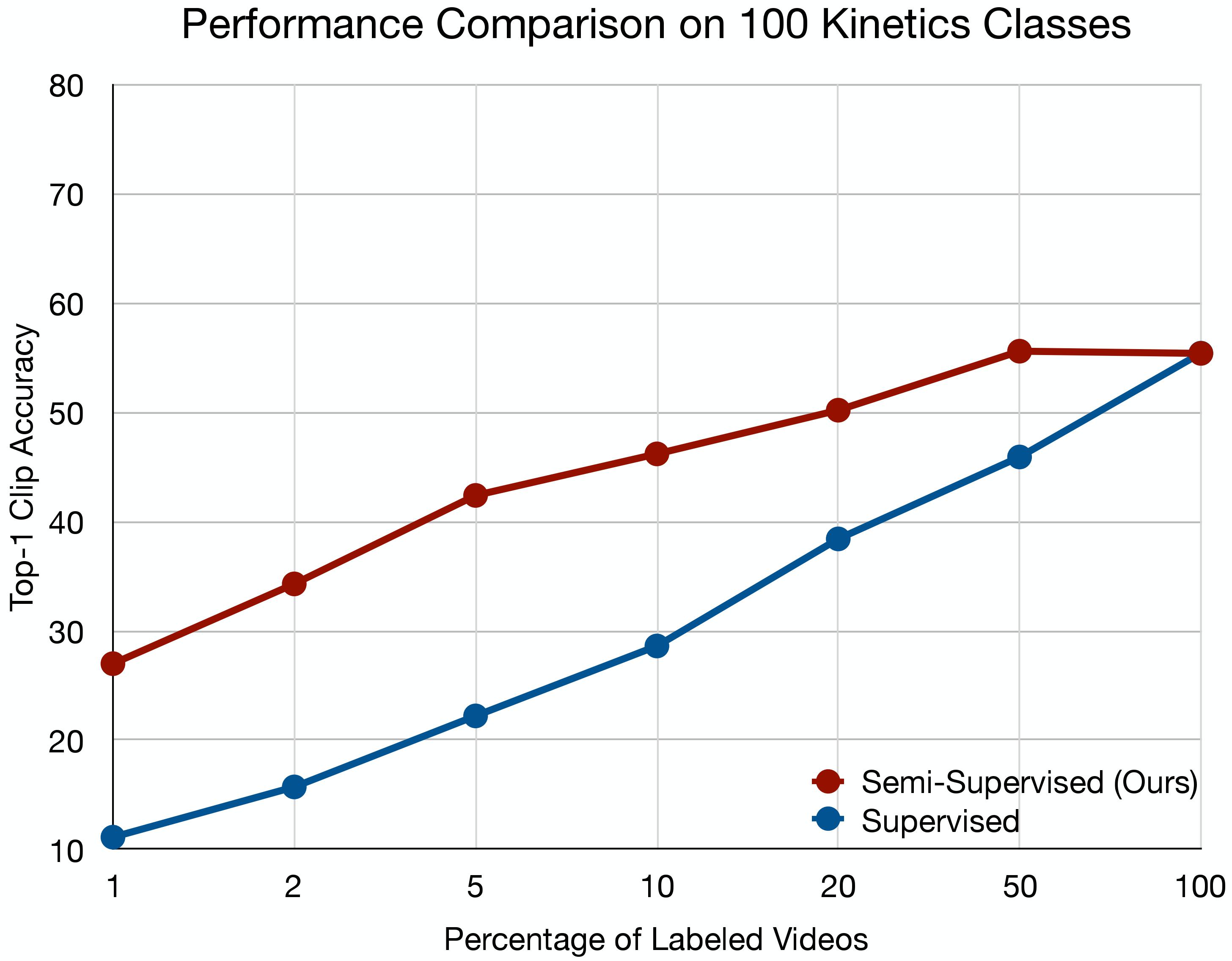}
\end{center}
\vspace{-0.5cm}
\caption{\small Video classification accuracy as a function of the fraction of labeled videos. With a small percentage of labeled examples, the 3D CNN trained by our proposed semi-supervised method significantly outperforms that trained in supervised setting.}
\label{fig:top1-kinetics100}
\vspace{-0.28cm}
\end{figure}

Similar to other machine learning problems, a large annotated dataset is critical for training CNNs (comprising millions of parameters) to achieve good performance for video classification. In spite of seemingly unlimited number of videos available on the internet, categorizing and curating these videos to create a useful video dataset such as~\cite{Kinetics, Something2Something, Sport-1M} is still expensive and tedious~\cite{oliver2018realistic}. The labels associated with the videos from social media are often noisy and need to be corrected manually. In addition, some videos require trimming as the action or video event often does not span through the video length~\cite{Kinetics}. 

In order to reduce the dependence on annotated datasets, several studies have investigated pretraining features with millions of web videos in a weakly supervised fashion where the video labels are noisy~\cite{LargeScaleWeakly, Girdhar_2019_ICCV, Sport-1M}. After feature learning, these methods finetune the overall network on the target dataset in a fully supervised fashion. Others have employed self-supervision for video feature learning~\cite{VideoDPC, shuffleandlearn}.

However, both finetuning (after pertaining) and self-supervised methods assume the existence of a high quality labeled dataset which incurs the aforementioned costs. A semi-supervised learning (SSL)  method, on the other hand, can reduce these costs by requiring fewer annotated training examples from target dataset. Several methods for semi-supervised learning in 2D image domain have reported very promising results~\cite{MeanTeacher,VAT}. A recent survey by~\cite{oliver2018realistic} compares the performances of these methods as well as suggests scenarios where SSL is a better choice than pretraining or self-supervised methods. Nonetheless, to the best of our knowledge, there has been no study that proposes an effective and robust semi-supervised algorithm for video classification with CNNs (see relevant works in Section~\ref{sec:related_work}).

In this paper, we propose a semi-supervised method, VideoSSL, for video classification with spatiotemporal networks. Given a small fraction of the annotated training samples, our proposed method leverages two supervisory signals  extracted from the unlabeled data to enhance classifier performance. As the first supervisory signal, we use  pseudo-labels~\cite{PseudoLabel} of the unlabeled data -- a technique that has been demonstrated to be highly effective on 2D images -- for semi-supervised learning of 3D video clips. We utilize the appearance cues of objects of interest, distilled by the prediction of a 2D image classifier CNN on a random video frame, as the second regularizer for VideoSSL.

\begin{figure}
\begin{center}
\includegraphics[width=0.5\textwidth]{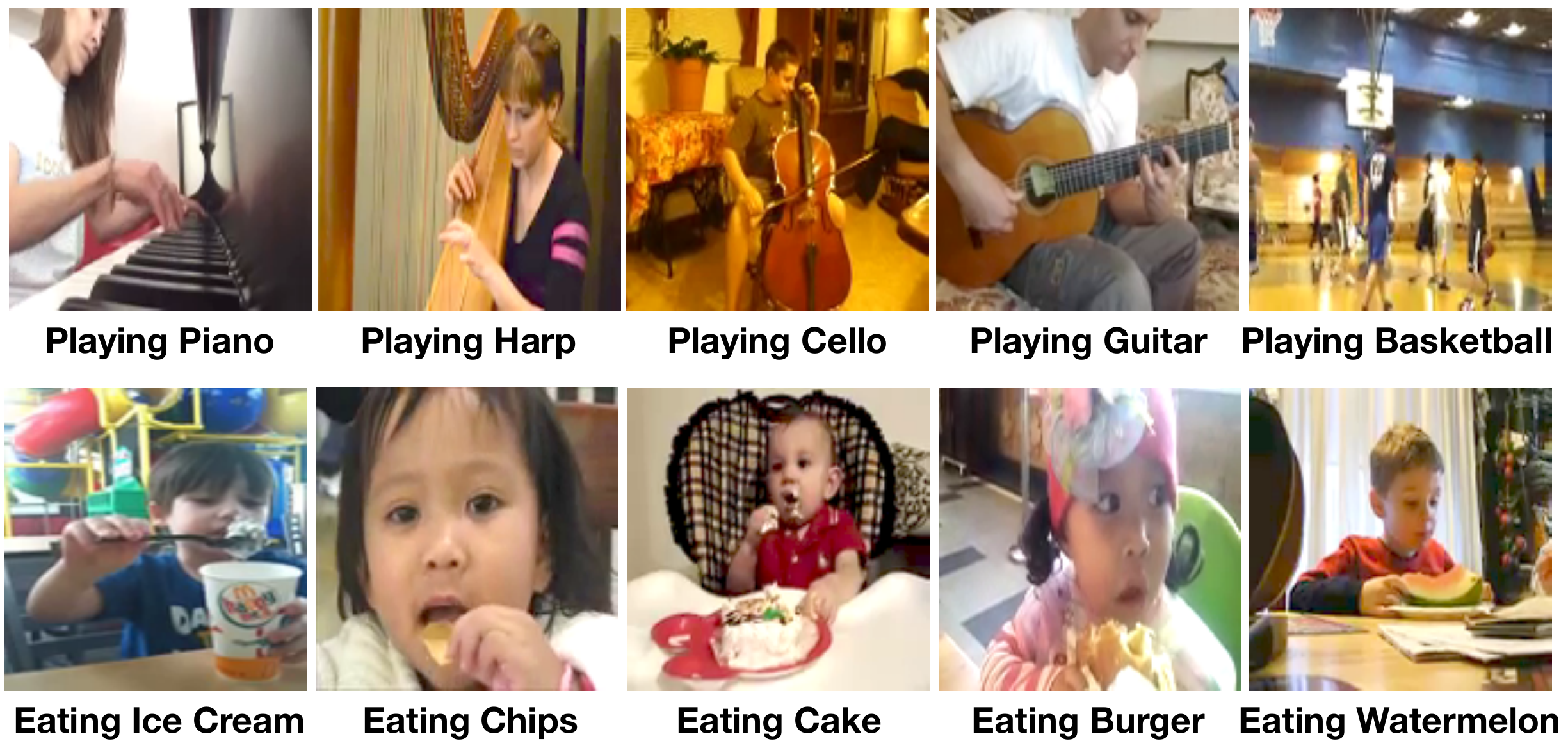}
\end{center}
\vspace{-0.5cm}
\caption{\small A single frame from some selected videos in Kinetics dataset. For these categories, object appearance in one single frame provides sufficient information to categorize them as playing instrument or sport (top row) and eating (bottom row) ~\cite{TimeCanTell, WhatMakesAVideo}.} 
\label{fig:noun}
\vspace{-0.35cm}
\end{figure}

Many, if not all, actions can be decomposed as one or more objects (noun) performing an activity (verb)~\cite{LargeScaleWeakly}. Consequently, a hint about the object (noun) appearance can offer a very strong indication of the actions being performed in the video clip~\cite{WhatMakesAVideo,TimeCanTell}; we  illustrate this insight with examples of actions in Figure~\ref{fig:noun}.  Girdhar \textit{et al.}~\cite{Girdhar_2019_ICCV} harnessed the appearance information in the form of the output probabilities of a 2D image classifier for \emph{pretraining} the spatiotemporal feature representation. Our algorithm proposes to use the predictions of 2D image classifiers as regulatory information for \emph{semi-supervised} training of 3D CNNs or their variants. In addition, we show that the capability of the video classifier can be further magnified by the incorporation of a semi-supervised technique.

We have tested our method on three most widely used datasets UCF101~\cite{UCF101}, HMDB51~\cite{HMDB51} and Kinetics~\cite{Kinetics}. On all the datasets, our proposed algorithm consistently trained 3D CNNs superior to those trained by the supervised algorithm from a small fraction of annotated examples.  The video classifiers learned by the proposed method can attain up to $20\%$ higher accuracy  than those of the classifiers trained by a fully supervised approach from limited data. Figure~\ref{fig:top1-kinetics100}  depicts a sample comparison between the performances of two networks trained by the proposed and supervised strategies. More interestingly, our proposed needed only $10\sim 20\%$ of the labeled data to produce a 3D CNN to match or supersede the accuracy of another network with the same architecture but trained from the whole dataset in a previous study~\cite{3DResNet}.  Our proposed technique can be generally applied to learn any 3D CNN variants for video classification.

%\textcolor{red}{REPHRASE/REWRITE --} We performed rigorous tests with different percentages of labeled examples and compared with different configurations and methods to test consistency -- the proposed method generated significant performance gains for each fraction of examples and against all combinations.

% -- an attempt that has not been undertaken before in computer vision community

This work contributes to the overall effort of video event recognition in multiple directions. We propose an accurate and robust semi-supervised training algorithm for 3D CNNs (or its variants) for video classification. We experimentally demonstrate that a straightforward execution of semi-supervised method does not yield a 3D video classifier with satisfactory performance. On the other hand, a calibrated utilization of the object appearance cues for semi-supervised learning profoundly improves the accuracy of the resulting model.  We validate the utility and consistency of our technique by reporting improved performances on different public datasets through rigorous testing under many different configurations.

%------------------------------------------------------------------
\section{Related Work}\label{sec:related_work}

\noindent \textbf{Video Classification:} Early studies on action recognition relied on hand designed features and models~\cite{niebles06bmvc,niebles10decomposable,laptev05interest, laptev08movies,liu09inthewild, wang09spatiotemporal}. Recently various networks have been proposed to capture both the spatial and temporal information for video classification tasks including: 2D CNN-based methods \cite{Sport-1M, Two-Stream, TSN}, RNN-based methods \cite{LRCN}, and 3D CNN-based methods \cite{C3D, P3D, I3D, 2Plus1D, LTC,gonda18bmvc}. In 2D CNN-based methods, high-level information are usually captured by a 2D CNN for each frame, and various fusion techniques including early and late fusion are applied to obtain the final prediction for each video \cite{Sport-1M, Two-Stream, TSN}. Some interesting analytical studies have recently investigated which categories of videos require temporal information for recognition \cite{TimeCanTell, WhatMakesAVideo}.

The 3D CNNs and their variants have made significant progress in video classification by simultaneously capturing spatial and temporal information \cite{I3D, C3D, P3D, 2Plus1D, gonda18bmvc, SlowFast}. However, due to the extra temporal dimension, the 3D CNNs usually have millions of parameters which may leads over-fitting when trained on small datasets. For that, in addition to learning from larger datasets like Kinetics~\cite{Kinetics},  there have been multiple efforts to pretrain the feature representations from millions of weakly annotated videos~\cite{Girdhar_2019_ICCV, LargeScaleWeakly}. 

\noindent \textbf{Semi-Supervised Learning:} Semi-supervised learning is a technique to train the network both with labeled and unlabeled data \cite{PI, PI2, VAT, MeanTeacher, PseudoLabel,oliver2018realistic}. Recently, several semi-supervised learning methods have been proposed for image classification. Considering the different random data augmentations to input data  and CNN configurations under dropout selection as noise to the learning process, \cite{PI, PI2} introduced a consistency loss between the network outputs from the same input sample at different training iterations, or their moving averages, as a regularization term for semi-supervised learning. In addition, Tarvainen and Valpola ~\cite{MeanTeacher} proposed to utilize a teacher model obtained from moving averages of past network weights to calculate a more \lq stable\rq~ prediction. {VAT is proposed by Miyato \textit{et al.}} to model the perturbations that added to the data which most significantly affect the output of the prediction function \cite{VAT}. {Grandvalet and Bengio suggests} minimizing entropy of the model predictions to generate more confident predictions~\cite{EntMin} whereas pseudo-label proposed to use the label predicted with highest confidence as the true label of the example for training~\cite{PseudoLabel}. Most of these methods have been tested on small dataset including CIFAR10 \cite{CIFAR10} and SVHN \cite{SVHN}, but their ability to adapt to large dataset has not been investigated yet. 

Semi-supervised learning of CNNs for 3D tasks has not yet received considerable interest in the community. A preliminary study by Zeng \textit{et al.}~\cite{zeng2017semi} employed an encode-decoder framework for action recognition but tested only on toy datasets containing few tens of images. The work of~\cite{Discrimnet} pretrains the feature representation through adversarial training and fine-tunes the discriminator on the target dataset; it does not learn the CNN in a semi-supervised manner. To the best of our knowledge, there are no works on semi-supervised learning of \emph{accurate} CNNs for video classification in computer vision literature before us. In our work, we experimentally demonstrate that the 2D semi-supervised learning techniques do not yield a satisfactory performance when directly extended to 3D network and therefore not useful. 

\noindent\textbf{Self-Supervised Learning: }Self-supervised learning is another trend of approach to learn visual features from unlabeled data \cite{VTS, SelfSurvey, shuffleandlearn,CubicPuzzles}. For learning video features from unlabeled videos, a network is trained to solve a pretext task and the label for pretext tasks are generated based on the attribute of the data. Various pretext tasks have been proposed to learn visual features from videos. Misra \textit{et al.}~\cite{shuffleandlearn} proposed to train network to verify whether the input frame sequence is in correct temporal order or not. Korbar \textit{et al.}~\cite{VTS} proposed to train network by verifying whether the input video segment and audio segment are temporally correspondent or not. A recent study by Zhai \textit{et al.}, combines the self-supervision with semi-supervised learning~\cite{SelfSemi}. However, this method was designed for and tested on 2D images only. 

\noindent\textbf{Knowledge Distillation:} Hinton \textit{et al.} \cite{Distil} originally proposed  to transfer the knowledge from several deep networks to one smaller network by optimizing the KL divergence of the distributions of the networks. Radosavovic \textit{et al.}~\cite{DataDistillation} proposed to distill knowledge from unlabeled data by using the prediction of a network whereas Garcia \textit{et al.}~\cite{RBGDDistill} propose to jointly transfer knowledge of different modalities to one modality.  

The work of~\cite{Girdhar_2019_ICCV} suggested  distilling  the appearance information of the objects of interest in the video through the output of a 2D image classification network for pretraining the 3D features of a video classifier. The 3D classifier is then finetuned on the target dataset using all its annotation. The proposed algorithm, on the other hand, uses the appearance information for semi-supervised training with a small fraction annotated samples from a dataset -- it does not require the target dataset to be exhaustively annotated. Such an approach could be beneficial for scenarios where collecting and annotating data is difficult and costly~\cite{oliver2018realistic}. 

\section{VideoSSL Training}\label{sec:videossl}

Our proposed algorithm VideoSSL trains a 3D CNN for video classification in a semi-supervised fashion. Motivated by the impressive performance of spatiotemporal 3D CNNs and their variants \cite{2Plus1D, P3D, I3D, 3DResNet, gonda18bmvc}, we used a 3D ResNet~\cite{3DResNet} that computes the (softmax) probabilities of different video classes.  It is worth pointing out that VideoSSL method can be used to learn any 3D CNN and its variants. In our semi-supervised setting, the softmax probabilities from a 2D image classifier are utilized as a supervisory or teaching signal to the training of 3D CNN. In the learning phase, the 3D CNN is designed to produce another output, which we also referred to as an embedding, with the same dimensionality as the 2D network output. 

\begin{figure*}
\begin{center}
\includegraphics[width=0.83\textwidth]{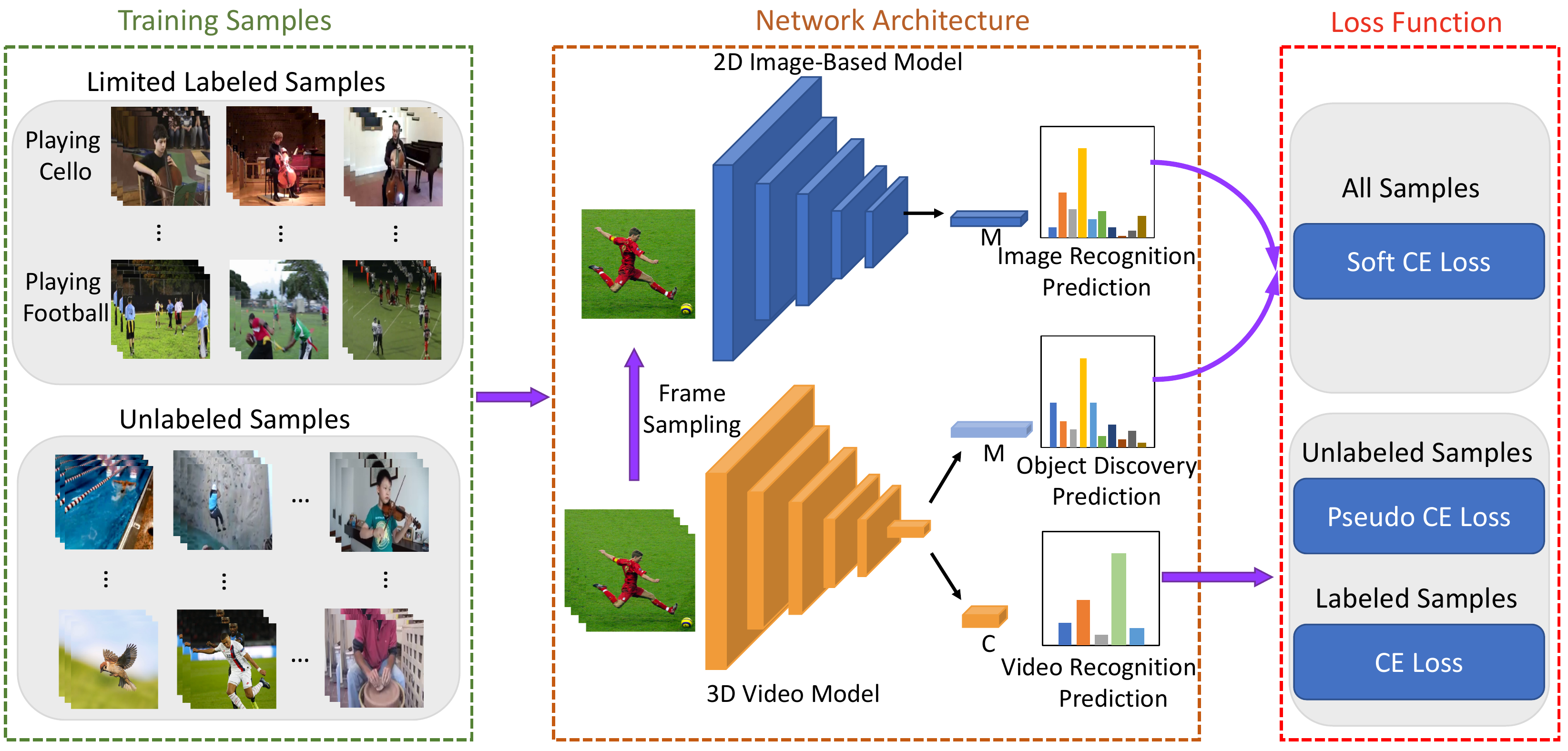}
\end{center}
\vspace{-0.25cm}
\caption{\small The framework of the proposed video semi-supervised learning approach. The 3D network is optimized with three loss functions: 1)~\textbf{CE Loss}: the video cross-entropy (CE) loss on the labeled data which paired with human-annotated labels, 2)~\textbf{Pseudo CE Loss}:  the pseudo cross-entropy loss on the pseudo-labels of unlabeled data, and 3)~\textbf{Soft CE Loss}: the soft cross-entropy loss on both unlabeled and labeled data to teach the video classification network to capture the appearance information.} 
\label{fig:Framework}
\vspace{-7pt}
\end{figure*}

The 3D CNN is trained by jointly minimizing three loss functions. The cross-entropy loss with respect to the labels of a small percentage of data points is backpropagated to update the weights of the 3D CNN. In addition, we also backpropagate the loss against the the pseudo-labels~\cite{PseudoLabel} computed by the 3D CNN on unlabeled examples. The third loss, which facilitates the knowledge distillation, is computed between the 2D image network prediction and the embedding from the 3D CNN computed for both labeled and unlabeled data. A schematic diagram of the whole training process is presented in Figure~\ref{fig:Framework} and we describe the losses used in VideoSSL in the following sections.

\subsection{Learning from Labeled Data}

Let $X = \{x_1, \dots, x_K\}$ denote the annotated video clips with corresponding category indicators $\{ y_1, \dots, y_K\}$ and $Z = \{ z_1, \dots, z_U \}$ be the unlabeled data in a batch of training examples. If there are $C$ video categories, i.e., $y_i \in \{0, 1\}^C$,  for any input video clip $x_i$, the 3D network produces a softmax probability $p(x_i) \in \mathbf{R}^C$ for $x_i$ to belong to any of the $C$ classes. Given the small set of examples $X$, the first loss 3D ResNet training minimizes is the cross-entropy loss.
\vspace{-0.2cm}
\begin{equation}\label{eqn:labeled}
    L_s = -  \sum_{x_i \in X}\sum_c ~y^c_{i} ~\log p^c (x_i)
\vspace{-0.2cm}
\end{equation} 
Here we omit the weight/parameter variables from the loss functions for better readability.

\subsection{Learning with Pseudo-Labels of Unlabeled Data}

Given a set of unlabeled examples $Z$, the method of pseudo-label computes an estimate of their true labels from the prediction of a classifier and use it to train the classifier itself~\cite{PseudoLabel}. In our proposed training, the label estimate $\hat{y}_i^c$ of $z_i$ for class $c$ is assigned to a pre-defined value $T$ if the prediction confidence $p^c(z_i)$ from 3D CNN on unlabeled sample $z_i$ exceeds $\delta$. A large $\delta$ enforces the algorithm to select highly confident samples; for such samples, predictions for less confidence classes become extremely small. As explained later (Section~\ref{sec:joint_loss}), we learn the network for a sufficient number of iterations before using its predictions for the pseudo-label approach. The resulting cross-entropy loss against the pseudo-labels can be formulated as follows. 
\vspace{-0.1cm}
\begin{equation}
    \hat{y}^c_i = \begin{cases} T, \quad  \text{if}~p^c(z_i) \ge \delta \\ p^c(z_i),~~\text{otherwise} \end{cases}
\end{equation}
% \hat{y}^c_i = \begin{cases} T, \quad  \text{if}~p^c(z_i) \ge \delta \\ p^c(z_i),~~\text{otherwise} \end{cases}
\begin{equation}
L_u = -  \sum_{z_i \in Z}\sum_c ~\hat{y}_i^c  ~\log p^c(z_i).  
\vspace{-0.1cm}
\end{equation}
We randomly select half of the examples in a batch from annotated examples and remaining half from examples without annotation. 

\subsection{Knowledge Distillation for All Data}\label{sec:distillation}

As several studies have already reported, appearance information can provide a strong cue for video/action recognition~\cite{WhatMakesAVideo, TimeCanTell, Girdhar_2019_ICCV}. Our method seeks to distill the information about the appearances of the objects of interest in the video by exploiting the softmax predictions of a 2D ResNet~\cite{ResidualNet} image classifier. The 2D ResNet we apply has already been trained on the ImageNet dataset~\cite{ImageNet} and its weights stay fixed throughout training and testing. For our VideoSSL approach, we distill the appearance information from both labeled and unlabeled video clips.

Given an image (or frame) $a$ from any video, Let us denote the output of the 2D ResNet as $h(a) \in \mathbf{R}^M$, where $M=1000$ for networks trained on ImageNet. In our experiments, we have randomly selected the frame $a$ from a video clip, both for training and testing. 

For each video $v \in \{ X \cup Z \}$, the 3D ResNet also produces another embedding $q(v) \in \mathbf{R}^M$ whose dimensionality matches that of the output of $h(a)$. During training we enforce the embedding from video classifier $q(v)$ to match the output of image classifier $h(a)$ when $a$ is a frame selected from $v$. The distillation loss utilized for this purpose is a soft cross-entropy loss that treats the 2D ResNet predictions as soft labels.
\vspace{-0.2cm}
\begin{equation}
    L_d = - \sum_{\substack{v \in \{ X \cup Z \}\\ a \in v}}\sum_{l=1}^M h^l(a) ~\log q^l(v)
\vspace{-0.2cm}
\end{equation}
We are using a knowledge distillation formula similar to that employed in~\cite{Girdhar_2019_ICCV}. However, as we explain in Section~\ref{sec:joint_loss}, our proposed VideoSSL method learns the overall 3D CNN (not just the features) by minimizing the distillation loss in conjunction with the supervised and pseudo-label losses in a semi-supervised fashion. This approach is fundamentally different from the feature learning of~\cite{Girdhar_2019_ICCV} for pretraining video classifiers.  

\subsection{Combined Loss Function}\label{sec:joint_loss} 

The overall training process trains the 3D network with a combined loss.
\vspace{-0.2cm}
\begin{equation}\label{eqn:joint_loss}
L = L_s + \lambda_u L_u + \lambda_d L_d.
\vspace{-0.2cm}
\end{equation}
The balancing weight for the pseudo-labels uses warm-up so that $\lambda_u = 1$ after a certain number of training iterations $\tau$. With a sufficiently large $\tau$, we can train the 3D CNN long enough to produce some meaningful predictions for pseudo-labels. The $\lambda_d = 1$ for all our experiments.

% 0$\exp{-5*(1- min())}$
% math.exp(-5 * (1 - min(iteration/args.warmup, 1))**2)

\section{Experimental Results}

In this section, we conduct extensive experiments to evaluate the proposed approach and compare with other semi-supervised learning methods from 2D image domain applied to video data. Our semi-supervised learning framework is trained and tested on several widely used datasets for video classification including: UCF101 \cite{UCF101}, HMDB51 \cite{HMDB51}, Kinetics \cite{Kinetics}. In the following, we first describe our experimental setting and network architecture \& training before reporting performances on these 3 datasets.

\subsection{Implementation Details}

We have used 3D ResNet-18~\cite{3DResNet} as a video classifier in all our experiments. This 3D ResNet architecture is very similar to the 2D ResNet~\cite{ResidualNet}, except all the convolutions are performed in 3D. That is, it has 4 convolutional blocks with different numbers of 3D convolutions (within the block) based on the ResNet size followed by an initial convolution and pooling. We have primarily experimented to 3D ResNet-18 (each block with two 3D convolutions) with 64, 128, 256, 512 feature maps. The 3D ResNet-18 has a $C$ class output for video categories. During training, it also produces a $M=1000$ length embedding for each video. The 2D ResNet-50 image classifier is collected from the pytorch repository. Our implementation was built around the code released by~\cite{3DResNet}.

The videos from all the datasets are resized to a spatial resolution at $136 \times 136$. During training, $ 16$  consecutive frames are randomly selected from each video as a training clip and a $112 \times 112$ patch is randomly cropped from each frame to form an input clip. The size of the input becomes $3$ channels $\times$ 16 frames $\times$ 112 $\times$ 112 pixels. The input to the 3D ResNet-18 was also normalized by the mean and variance of the sport-1M dataset. We used random crop and temporal jittering for data augmentation in all our experiments. The input size and data preprocessing strategies are very similar to existing studies~\cite{2Plus1D,Girdhar_2019_ICCV,LargeScaleWeakly}.

All the models are trained on different percentage of labeled data. We have randomly selected different percentages $P$ of labeled examples from each of the datasets, e.g., $P \in \{ 5, 10, 20 ,50\}$.  In our VideoSSL training, we used $P$ percentage of labeled data to compute the supervised loss in Equation~\ref{eqn:labeled}. Annotations for all remaining examples were ignored in the semi-supervised setting and treated as unlabeled examples. Given the split of annotated and unannotated examples, our VideoSSL learning minimizes the joint loss in Equation~\ref{eqn:joint_loss} to learn  a 3D CNN from scratch. For all the experiments on the same dataset, the same testing splits are used for fair comparison.

We have used the Stochastic Gradient Descent (SGD) with momentum $0.9$ and weight decay $0.001$ as a minimizer for the joint loss. The initial learning rate during learning was set to 0.01 and was decreased by a factor of 10 every 40000 iterations. The batch size for every optimization step was 128 distributed among multiple GPUs. For pseudo-label technique,  $T$ and $\tau$ were set to $10$ and $2 \over 3$ of the total iterations respectively. 

For all the experiments below, we report the Top-1 clip and video accuracy values on the validation or test datasets.  After training, the prediction of the 3D ResNet on the center video clip (both spatial and temporal) is reported as the  clip Top-1 accuracy. The video accuracy is the average of the classifier confidences on all consecutive non-overlapping clips within the video.
\begin{table*}[!t]
\vspace{-0.3cm}
\caption{\small The performance comparison on UCF-101 dataset. All values reported are Top-1 accuracy values. The proposed method consistently generates the most accurate CNNs.}
\vspace{-0.6cm}
\begin{center}
\begin{tabular}{l|c c|c c|c c|c c|c c|c c|c c}
\hline
\multirow{2}{*}{\%Label} &  \multicolumn{2}{c|}{Supervised\cite{3DResNet}} & \multicolumn{2}{c|}{PL\cite{PseudoLabel}}&\multicolumn{2}{c|}{MT\cite{MeanTeacher}}&\multicolumn{2}{c|}{SD}&\multicolumn{2}{c|}{MT+SD}&\multicolumn{2}{c|}{S\textsuperscript{4}L \cite{SelfSemi}}&\multicolumn{2}{c}{Ours}  \\
\cline{2-15}
& clip & video& clip &video& clip &video& clip &video& clip &video& clip &video& clip &video \\
\hline
5 & 15.1 & 16.9 & 17.2 & 17.6 & 15.3 & 17.5 & 29.3 & 31.2 & 28.4 & 30.3 & 21.0 & 22.7 & 30.9 & \textbf{32.4} \\
10 & 21.6 & 24.0 & 23.5 & 24.7 & 24.0 & 25.6 & 38.6 & 40.7 & 37.5 & 40.5 & 27.1 & 29.1 & 40.2 & \textbf{42.0}\\
20 & 30.0 & 32.2 & 33.9 & 37.0 & 33.4 & 36.3 & 42.1 & 45.4 & 41.7 & 45.5 & 34.7 & 37.7 & 46.2 & \textbf{48.7} \\
50 & 35.1 & 38.3 & 43.9 & 47.5 & 42.5 & 45.8 & 49.8 & 53.9 & 49.2 & 53.0 & 44.9 & 47.9 & 51.5 & \textbf{54.3} \\
\hline
\end{tabular}
\end{center}
\label{tab:res-ucf101}
\vspace{-0.3cm}
\end{table*}

\begin{table*}[!t]
\caption{\small The performance comparison on Kinetics-100 dataset. The proposed method consistently improves both the clip and video Top-1 classification accuracy and outperforms all other methods.}
\vspace{-0.6cm}
\begin{center}
\begin{tabular}{l|c c|c c|c c|c c|c c|c c|c c}
\hline
\multirow{2}{*}{\%Label} & \multicolumn{2}{c|}{Supervised\cite{3DResNet}} & \multicolumn{2}{c|}{PL\cite{PseudoLabel}}&\multicolumn{2}{c|}{MT\cite{MeanTeacher}}&\multicolumn{2}{c|}{SD}&\multicolumn{2}{c|}{MT+SD}&\multicolumn{2}{c|}{S\textsuperscript{4}L \cite{SelfSemi}}&\multicolumn{2}{c}{Ours}  \\
\cline{2-15}
& clip & video& clip &video& clip &video& clip &video& clip &video& clip &video& clip&video \\
\hline
5 & 23.6 & 27.2 & 24.8 & 27.8 & 23.8 & 27.8 & 40.2 & 45.2 & 40.8 & 46.6 & 29.6 & 33.0 & 43.1 & \textbf{47.6}  \\
10 & 31.2 & 36.3 & 34.6 & 38.9 & 31.5 & 36.4 & 44.7 & 49.8 & 43.9 & 49.4 & 37.5 & 43.3 & 48.4 & \textbf{52.6} \\
20 & 40.7 & 46.8 & 41.8 & 48.0 & 40.8 & 47.1 & 49.8 & 55.6 & 50.0 & 55.3 & 44.7 & 51.1 & 51.3 & \textbf{57.7} \\
50 & 49.6 & 55.5 & 51.2 & 59.0 & 51.2 & 59.3 & 57.3 & 63.8 & 57.6 & 63.9 & 49.1 & 54.6 & 58.2 & \textbf{65.0} \\
\hline
\end{tabular}
\end{center}
\label{tab:res-kinetics100}
\vspace{-15pt}
\end{table*}

% \begin{table*}[!t]
% \caption{\small The performance comparison on Kinetics-100 dataset. The proposed method consistently improves both the clip and video Top-1 classification accuracy and outperform all other methods.}
% \vspace{-0.6cm}
% \begin{center}
% \begin{tabular}{l|c c|c c|c c|c c|c c|c c}
% \hline
% \multirow{2}{*}{\%Label} & \multicolumn{2}{c|}{Supervised\cite{3DResNet}} & \multicolumn{2}{c|}{PL\cite{PseudoLabel}}&\multicolumn{2}{c|}{MT\cite{MeanTeacher}}&\multicolumn{2}{c|}{SD}&\multicolumn{2}{c|}{MT+SD}&\multicolumn{2}{c}{Ours}  \\
% \cline{2-13}
% & clip & video& clip &video& clip &video& clip &video& clip &video& clip &video \\
% \hline
% 5 & 23.6 & 27.2 & 24.8 & 27.8 & 23.8 & 27.8 & 40.2 & 45.2 & 40.8 & 46.6 & 43.1 & \textbf{47.6}  \\
% 10 & 31.2 & 36.3 & 34.6 & 38.9 & 31.5 & 36.4 & 44.7 & 49.8 & 43.9 & 49.4 & 48.4 & \textbf{52.6} \\
% 20 & 40.7 & 46.8 & 41.8 & 48.0 & 40.8 & 47.1 & 49.8 & 55.6 & 50.0 & 55.3 & 51.3 & \textbf{57.7} \\
% 50 & 49.6 & 55.5 & 51.2 & 59.0 & 51.2 & 59.3 & 57.3 & 63.8 & 57.6 & 63.9 & 58.2 & \textbf{65.0} \\
% \hline
% \end{tabular}
% \end{center}
% \label{tab:res-kinetics100}
% \end{table*}

\subsection{Baseline Methods}

In all our experiments on different datasets, we have compared the performance of the CNN trained by the proposed algorithm to those trained by different methods as well as their combinations listed below. Unless otherwise mentioned, the same experimental setup was maintained for all the experiments.
\begin{enumerate}
\itemsep-0.5em 
    \item Supervised baseline (Supervised) learns the 3D Resnet18~\cite{3DResNet} from only the labeled examples.
    \item MeanTeacher (MT) applies the method of ~\cite{MeanTeacher} on video data.
    \item PseudoLabel (PL) applies the technique of~\cite{PseudoLabel} on video data.
    \item Supervised with Distillation  (SD) uses the knowledge distillation loss, as described in Section~\ref{sec:distillation}, along with the supervised loss for the training.
    \item The self-supervised and semi-supervised learning method (S\textsuperscript{4}L) of ~\cite{SelfSemi} extended to video data. We adopt the S\textsuperscript{4}L-rotate strategy originally proposed for 2D images for 3D videos. In particular, we minimize the cross-entropy loss on labels and rotations of the annotated and unlabeled videos respectively in our experiments. 
% S\textsuperscript{4}L~\cite{SelfSemi} was originally proposed to boost the performance of image semi-supervised learning method by incorporating the self-supervision signal as regularization. We adapt this method to the video semi-supervised learning. Specifically, the labeled video are trained with both the cross-entropy loss for the video classification and the clip rotation cross-entropy loss for the rotation recognition task, while the unlabeled videos are used to train the network by minimizing the rotation recognition loss for the clip rotation recognition task.
\end{enumerate}

For supervised learning, we used only the labeled examples, as given by the percentage $P$, to train the CNN from scratch.

\subsubsection{Results on UCF101 Dataset}
\noindent \textbf{Dataset:} UCF101 is a widely used dataset for human action recognition \cite{UCF101}. It consists of $13,320$ videos belong to $101$ action classes and contains approximately $130$ videos for each class. Although relatively small in size, it is a balanced dataset and each class has around  $100$ videos for training. Videos have the spatial resolution of $240 \time 320$ pixels and $25$ FPS frame rate. There are three training/testing splits available for this dataset, and the split $1$ is used for all the experiments in our paper.

\noindent \textbf{Performance Comparison:} Table~\ref{tab:res-ucf101} shows the clip and video Top-1 accuracy of our proposed method and the baselines for video classification with 3D ResNet-18. As shown in the table,  our proposed strategy amplifies the video Top-1 accuracy of the 3D ResNet-18 by more than $16\%$ with $\{5\%, 10\%, 20\%, 50\%\}$ annotated samples. Across all percentages of labeled data, our algorithm produces the most accurate classifier among all other techniques. 

These experiments also suggest that the straightforward application of the existing semi-supervised methods PL~\cite{PseudoLabel} and MT~\cite{MeanTeacher} to 3D video classifier is not beneficial. It is interesting to observe that the accuracy of MT is similar or worse than PL, which contrasts the findings of~\cite{oliver2018realistic} albeit for 2D images. However, as~\cite{SelfSemi} points out, such an outcome has been observed in practice before. The adaptation of knowledge distillation~\cite{Girdhar_2019_ICCV} is instrumental in achieving good performances for semi-supervised learning from a limited percentage of data.  The combination of the semi-supervised PL technique to knowledge distillation further improves the accuracy of the resulting 3D CNN by contributing additional information to the training process. 

Perhaps the most compelling outcome of our experiments is, with \emph{only $10\%$ of annotated data} the proposed method can achieve the same video Top-1 accuracy of the 3D ResNet-18 trained from scratch in a fully supervised manner in~\cite{3DResNet}. With \emph{$50\%$ labeled examples} the proposed approach produces a $12\%$ more accurate CNN.

\begin{table*}[!t]
\vspace{-0.3cm}
\caption{The performance comparison on HMDB51 dataset. All values reported are Top-1 accuracy values. }
%{\color{blue} For the final result, lamda = 1/1.2, iterations = 8k, learning rate 4k.}
\vspace{-0.6cm}
\begin{center}
\begin{tabular}{l|c c|c c|c c|c c|c c|c c|c c}
\hline
\multirow{2}{*}{\%Label} &  \multicolumn{2}{c|}{Supervised\cite{3DResNet}} & \multicolumn{2}{c|}{PL\cite{PseudoLabel}}&\multicolumn{2}{c|}{MT\cite{MeanTeacher}}&\multicolumn{2}{c|}{SD}&\multicolumn{2}{c|}{MT+SD}&\multicolumn{2}{c|}{S\textsuperscript{4}L \cite{SelfSemi}}&\multicolumn{2}{c}{Ours}  \\
\cline{2-15}
& clip & video& clip &video& clip &video& clip &video& clip &video& clip &video& clip &video \\
\hline

{40} &17.1   & 18.0  & 26.3 & 27.3 & 26.4  & 27.2 & 31.6 & 32.6 & 32.1 & 32.3 & 28.8 & 29.8 & 32.6 & 32.7 \\

{50} & 29.1  & 30.7  & 30.9 & 32.4 & 29.2  & 30.4 & 34.1 & 35.1 & 30.8 & 33.6 & 28.9 & 31.0 & 34.9 & \textbf{36.2} \\

%\color{blue}50 & 29.1  & 30.7  & --  & --  & --  & -- & \color{blue}35.0 & \color{blue}36.2   & -- & -- & -- & -- & \color{blue}35.9 & \color{blue}37.3 [1:1.2]\\

{60} & 30.0  & 31.2  & 31.4  & 33.5 & 31.1  & 32.2   & 35.4 & 36.3  & 34.5 & 35.7 & 32.5 & 35.6 & 35.7 & \textbf{37.0} \\

%{\color{blue}60} & --  & --  & --  & & --  & --  & 36.0 & 38.3 & -- & -- & -- & -- & 36.9 & 37.5 [1:1.2]\\

\hline
\end{tabular}
\end{center}
\vspace{-0.2cm}
\label{tab:res-hmdb51}
\end{table*}

\subsubsection{Results on Kinetics Dataset}

\noindent\textbf{Dataset:} Kinetics is a large-scale dataset for video understanding tasks \cite{Kinetics}. The Kinetics-400 version, provides $306,245$ 10-second videos belong to $400$ action classes. Since many videos are not available on the YouTube any more, we were able to download $226,127$ and $18,613$ videos for training and validation respectively. This dataset is significantly larger than other popular video datasets and has been increasingly popular in the action recognition community~\cite{LargeScaleWeakly,3DResNet,Girdhar_2019_ICCV, SlowFast}.

However, the distribution of videos across different categories is not balanced in Kinetics-400. Some classes in this dataset contain over $900$ videos whereas more than $80$ classes contain less than $300$ videos. In order to create a more homogeneous distribution, {we have picked 100 classes with an at least $700$ training videos in each category.} This subset of Kinetics dataset is referred to as Kinetics-100 in this paper.

% \textcolor{red}{**Longlong: explain how we picked the 100 classes and how many train/test videos in the Kinetics 100 class**}
% For the semi-supervised learning methods, most of the methods are only evaluated on the small dataset such as CIFAR10 and STL10, while only a few of them have been evaluated on large dataset. These methods perform well on small dataset may not be able to scale to large dataset. To evaluate the performance of our methods, we conduct realistic evaluation of our model on the imbalanced public dataset Kinetics400 dataset. More than 10 categories have only around 200 videos in our dataset, while some category have around 1000 videos. When only 1\% video have labels, some category only have 2 labeled videos from training. The experiments results on Kinetics400 dataset is shown in Table.~\ref{tab:res-400}.

\noindent\textbf{Performance Comparison:} As shown in Table.~\ref{tab:res-kinetics100}, our method consistently improve the accuracy of the 3D ResNet over that trained by the supervised method by a significant amount. The improvement over the supervised method reduces from roughly $20\%$ to $10\%$ in video Top-1 accuracy when the labeled data increases from $5\%$ to $50\%$. It is expected that the difference in accuracy between semi and fully supervised methods will decrease with the increase of labeled data. The results suggest that the off the shelf application of the existing semi-supervised methods (PL and MT) offer little benefit to video classification of Kinetics dataset as well.

% \textcolor{red}{Remove:We are not sure why the performance of MT+SD is worse than SD only for $10$ and $20\%$ labels.}

\emph{The proposed method can achieve a higher video Top-1 accuracy of the 3D ResNet-18 trained by fully supervised training in~\cite{3DResNet} with only $20\%$ of annotated data in the Kinetics dataset}. Although we are using a quarter of the total classes evaluated in~\cite{3DResNet}, this result on a more realistic Kinetics dataset sheds light on the strength of the proposed technique for semi-supervised video classification.

\begin{figure}
\begin{center}
\includegraphics[width=0.45\textwidth, height=0.3\textwidth]{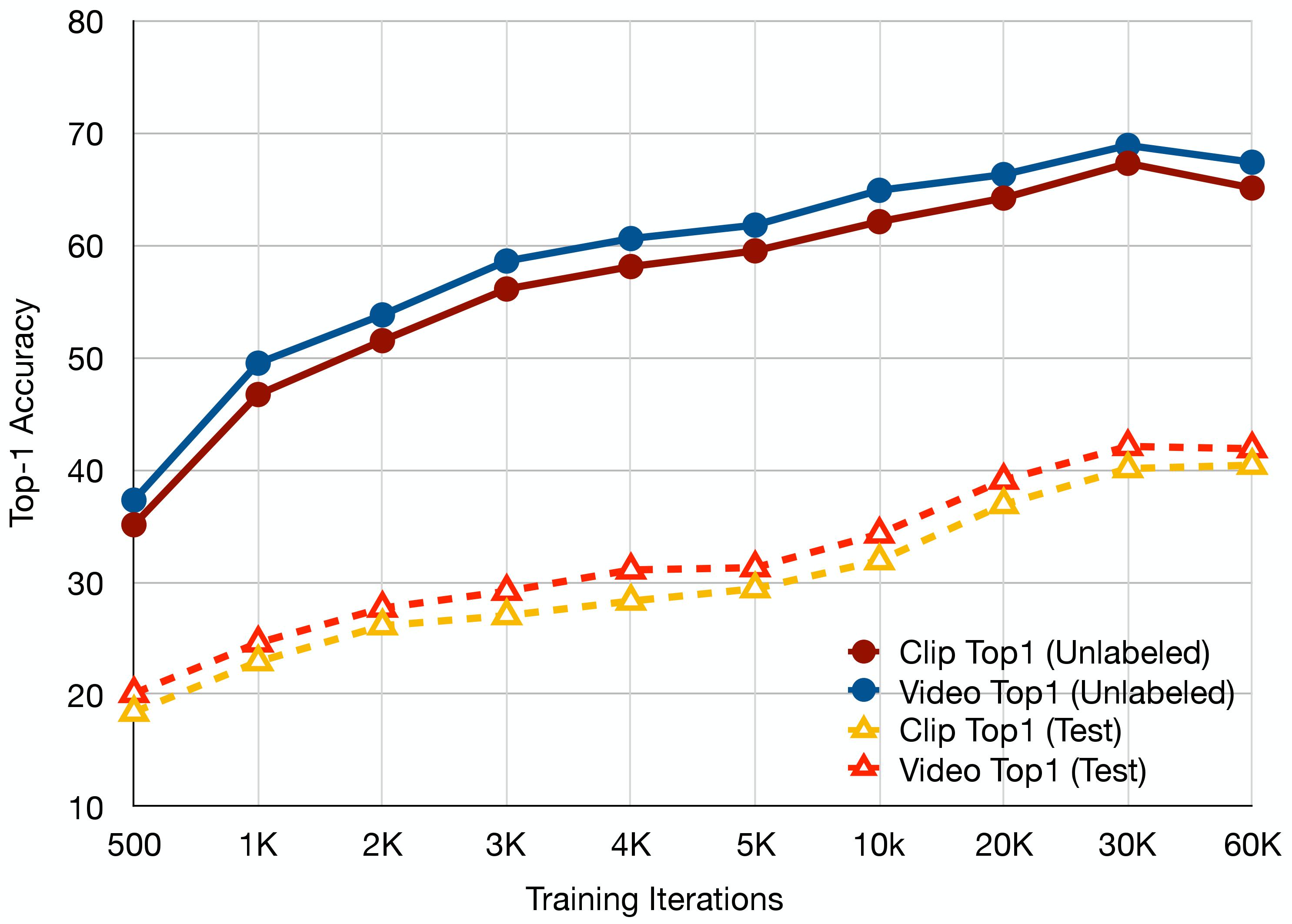}
\end{center}
\vspace{-0.4cm}
\caption{\small Progression of clip and video Top-1 accuracy of 3DCNN trained by the proposed algorithm on the unlabeled training samples (solid lines) and test split (dashed lines) of UCF101 dataset with training iterations.} 
\label{fig:AccIterations}
\vspace{-0.4cm}
\end{figure}

\begin{figure}[h]
\begin{center}
\includegraphics[width=0.5\textwidth]{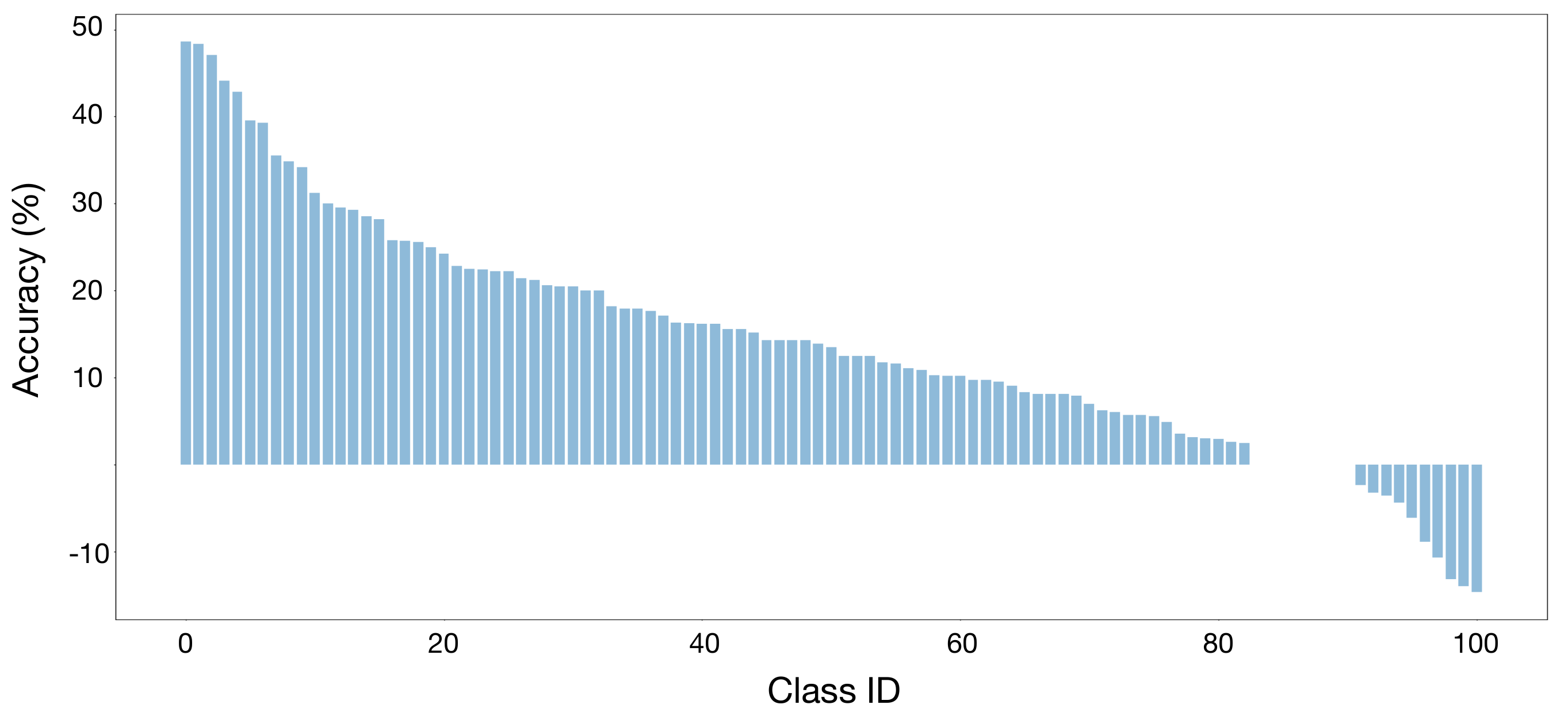}
\end{center}
\vspace{-0.5cm}
\caption{\small Per-class accuracy improvements on UCF101 dataset between the network trained by the baseline supervised and our proposed method. The y-axis plots the improvements gained for classes shown on x-axis. For $90\%$ of the classes the proposed algorithm can boost the performance of the learned model.} 
\label{fig:improvement_per_class}
\vspace{-0.2cm}
\end{figure}

\begin{figure*}[h]
\begin{center}
\includegraphics[width=\textwidth]{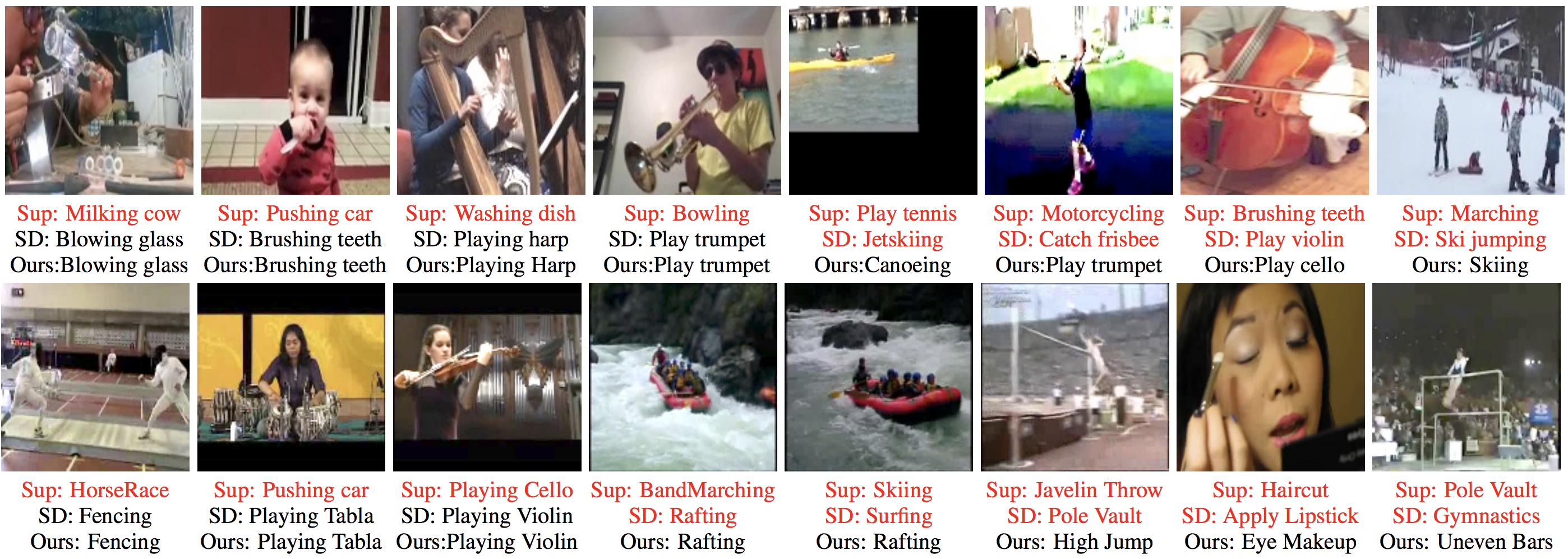}
\end{center}
\vspace{-0.5cm}
\caption{\small Qualitative comparison of our algorithm with baseline methods, top row: Kinetics100, bottom row: UCF101. Each image is a frame from a video that was correctly classified by the 3D CNN learned by the proposed method from $10\%$ examples. Sup, SD, and ours refer to the predictions of the supervised~\cite{3DResNet}, SD and proposed method respectively. The detections from supervised CNN appear to be arbitrary compared to the video category whereas those from SD seem to capture and exploit the scene charateristics.} 
\label{fig:qual}
\vspace{-0.5cm}
\end{figure*}

\subsubsection{Results on HMDB51 Dataset}

\noindent\textbf{Dataset:} HMDB51 is another  widely used dataset for human action recognition \cite{HMDB51}. It consists of $6,770$ videos belong to $51$ action classes and each class has roughly $70$ videos for training. There are three splits available for this dataset and we used split $1$ for all our experiments. In spite of the smaller size compared to UCF101 and Kinetics, the performances of the existing techniques have been lower than those on the other two datasets~\cite{3DResNet,Distil, C3D}. This implies a higher complexity to deal with HMDB51 with respect to the other datasets. 

% {\color{red}Since this dataset is too small, both the training and validation split is used during training.}

\noindent\textbf{Performance Comparison:} Due to the relatively small size of HMDB51 dataset, the performances of our proposed method compared against the baseline methods on $\{40\%, 50\%, 60\%\}$ annotated examples instead. 

Table~\ref{tab:res-hmdb51} compares the performances of the proposed algorithm and the baseline methods. The findings from this experiment conform almost exactly to those from the UCF101 and Kinetics -- our VideoSSL trained 3D CNNs from different percentages of annotations that are consistently superior to those trained by the supervised, exiting semi-supervised and also the self semi-supervised techniques. Likewise, our approach produced a 3D ResNet-18 more accurate than that trained by~\cite{3DResNet} with only $50\%$ of annotations.

% [5%]  clip top1: 0.035948 clip top5: 0.416993  vid top1: 0.038562 vid top5: 0.436601 
% [10%] clip top1: 0.077778 clip top5: 0.498693  vid top1: 0.071242 vid top5: 0.519608
% [20%] clip top1: 0.165359 clip top5: 0.524837  vid top1: 0.166667 vid top5: 0.537908 
% [50%] clip top1: 0.362745 clip top5: 0.649020  vid top1: 0.373203 vid top5: 0.660784 

% As shown in the table, our methods can significantly improve the performance when the training data is extremely limited. When only a few videos have labels (less than 10), our methods can significantly improve the clips and video TOP-1 accuracy by almost $20$\%. 

\subsection{Analysis of Training}
The success of a semi-supervised method relies heavily on how well it learns to classify the unlabeled samples during the training process. In Figure~\ref{fig:AccIterations}, we plot the  accuracy progression of the CNN under training on the unlabeled training data as well as the test data data  at different training iterations. This experiment was performed on $10\%$ labeled examples of UCF101 dataset. The plot clearly illustrates how the performance of the CNN was improved by the proposed method over the training process on both clip and video classifications.

% \begin{table}[ht]
% \caption{Confidence Analysis on UCF101 Dataset}
% \begin{center}
% \begin{tabular}{l|c|c|c|c|c}
% \hline
%  \%Label  & Videos & # \> 0.95      & accuracy &Overall \\
% \hline\hline
% 5        & 9060    & 3293   & 2976   & 90.3 \\
% 10       & 8583    & 5434   & 4885   & 89.9 \\
% 20       & 7629    & 5954   & 5677   & 95.2 \\
% 50       & 4768    & 4368   & 4308   & 94.7 \\
% \hline
% \end{tabular}
% \end{center}
% \label{tab:confidence_analysis}
% \end{table}

% {\color{red}
% Pick some samples that ImageNet and supervised make wrong decision while our method is right. Not only classes, we have to find a good video and pick a frame that supports our claim.

% LONGLONG and ZHE: READ THE FOLLOWING PARAGRAPHS TO GET A SENSE OF THE CATEGORIES THAT WOULD BE USEFUL
% Possible categories: \{bench press, push up\}, \{pole vault, javelin throw\}, \{rowing, rafting\}, \{playing cello, playing violin\}
% }

To investigate the comparative strength of the proposed method over the supervised training across different categories, we compute the per-class relative improvement of achieved by our strategy. Figure~\ref{fig:improvement_per_class} plots the category-wise increase the classification accuracy (clip Top-1) of the network trained by our method compared to that trained by the supervised approach with $10\%$ labels of UCF101. As seen on the plot, the performance of the 3D ResNet learned by our method improved for $90\%$ of the categories. Example classes such as Boxing Speed Bag (+48.6), Playing Tabla (+48.4), Fencing (+47.1), Sumo Wrestling (44.1), Rafting(+42.3), Bench Press (+39.6), Playing Violin (+39.3), Drumming (+35.5), Band Marching (+34.9), Biking (+34.2) imply the appearance information of the objects of interest in the video played a major role in this improvement. 

There are categories in both the UCF101 and Kinetics100 datasets where the proposed VideoSSL trained better classifiers (with $10\%$ labels) than the SD method that partially utilizes the object appearance cues. Figure~\ref{fig:qual} shows some representative frames from these classes from both these datasets. As can be expected, the misclassification of the supervised method~\cite{3DResNet} appears to be rather arbitrary with respect to the actual categories. SD, on the other hand, classifies these video into  (wrong) categories with very similar scene characteristics. Examples of SD misclassification predict throwing frisbee for passing football or jetskiing for canoeing.  The proposed technique utilized additional knowledge supplied by the pseudo-label method to resolve the confusion and achieve a superior performance on these categories.
\begin{figure}
\begin{center}
\includegraphics[width=0.35\textwidth, height=0.2\textwidth]{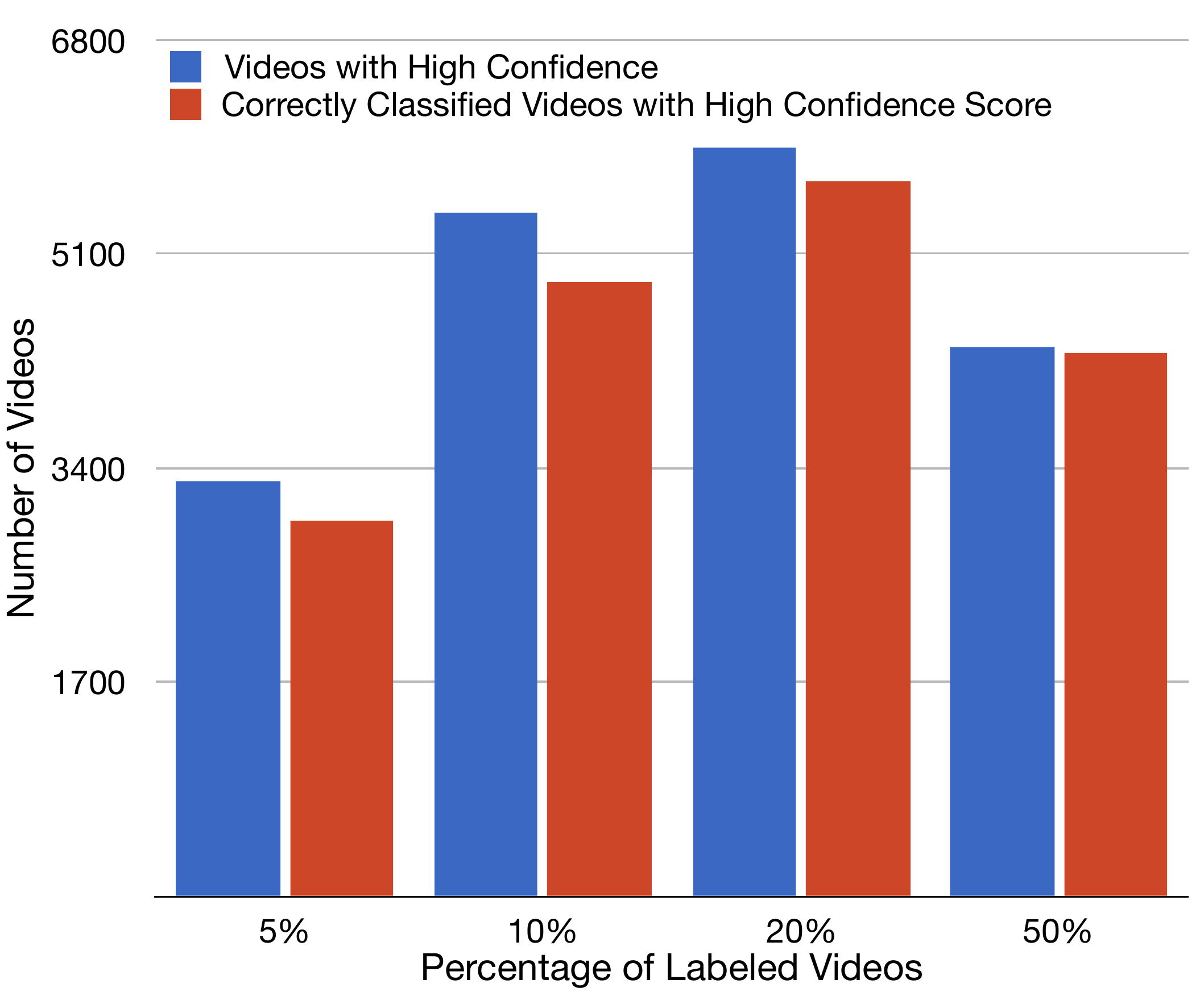}
\end{center}
\vspace{-0.6cm}
\caption{\small Number of unlabeled videos in the training set correctly classified by our training strategy. Blue: unlabeled videos receiving a prediction with high confidence $p^c(z) > 0.95$ for some $c$; Red: out of the videos with high confidence, how many were classified correctly.} 
\label{fig:PL-Analysis}
\vspace{-0.5cm}
\end{figure}

In Figure~\ref{fig:PL-Analysis}, we plot the number of videos that receive a high confidence score (blue) and how many of those are correctly classified by the CNN being trained. It shows that more than $90$\% of the videos receiving confident predictions ($p^c(z)> 0.95$) are correctly classified by CNNs trained by our approach with different fractions of annotations. This observation reaffirms our claim of the capability and robustness of our algorithm and suggests potential use of the resulting classifier for frameworks to expand  datasets through active learning. 

\section{Conclusion}

This study conducts the first comparative study and proposes a new algorithm for semi-supervised learning of video classifier. We show in this work that a straightforward application of the existing semi-supervised methods (that are originally developed for 2D images) cannot achieve satisfactory performance for 3D video classification. The proposed method exploits the appearance information of the object of interest in video to produce highly accurate 3D classifiers given limited annotated examples. From only $20\sim50\%$ annotated samples, the proposed approach can learn CNNs that can potentially outperform those trained in a fully supervised manner. We have tested the accuracy and robustness of our algorithm on three most widely used datasets with different percentages of training labels and compared against the several baseline combinations. We hope that our proposed learning strategy will be useful for reducing the costs for creating a training dataset for video understanding and will instigate more efforts on semi-supervised video training.

% \vspace{3mm}

\noindent\textbf{Acknowledgement.}
This material is partially based upon the work supported by National Science Foundation (NSF) under award number IIS-1400802.

{\small
\bibliographystyle{ieee_fullname}
\bibliography{VideoSSL}
}

\end{document}